\documentclass[final,5p,times,twocolumn]{elsarticle}

\usepackage{amsmath}
\usepackage{amssymb}
\usepackage{booktabs}
\usepackage[final]{microtype}
\usepackage{lipsum}
\usepackage[section]{placeins}
\usepackage[T1]{fontenc}

\begin{document}

\begin{frontmatter}

\title{Class-Specific Branch Attention for Mitigating 
Gradient Interference under Class Imbalance}

\author[inst1]{Arush Singhal}
\author[inst2]{Dr. Umang Soni}

\address[inst1]{Thapar Institute of Engineering and Technology, Patiala, India}
\address[inst2]{Netaji Subhash University of Technology, New Delhi, India}

\begin{abstract}
Deep neural networks trained under severe 
class imbalance often exhibit degraded performance, 
typically attributed to statistical bias. In this work, 
we identify a complementary optimization-level pathology: 
inter-class gradient interference within shared 
representations, where gradients from majority classes 
suppress minority-class learning. To analyze this 
phenomenon, we introduce a diagnostic framework based 
on layer-wise gradient flow analysis and a Gradient 
Conflict Matrix, which quantifies interference using 
cosine similarity between class-specific gradients. 
Using this framework, we study multi-branch convolutional 
architectures and propose a lightweight modification, 
Class-Specific Branch Attention (CSBA), that enables 
branch-specific channel reweighting to reduce gradient 
coupling. This mechanism promotes implicit feature 
decoupling across branches while preserving architectural 
simplicity. Empirically, CSBA improves minority-class 
performance, increasing the F1 score for the 
Physical-Damage class from $0.261$ to $0.522$ under 
severe imbalance, while maintaining comparable overall 
accuracy. These results indicate that mitigating gradient 
interference can directly enhance minority-class 
representation. Validation on CIFAR-10-LT confirms that 
this behavior generalizes across imbalanced visual 
recognition settings, with Macro-F1 improving from 
$0.595$ to $0.655$. More broadly, our findings highlight 
the importance of considering optimization dynamics 
alongside statistical methods when designing architectures 
for imbalanced learning.
\end{abstract}

\begin{keyword}
Class imbalance \sep Gradient interference \sep 
Multi-branch neural networks \sep Attention mechanisms \sep 
Deep learning optimization \sep Imbalanced visual recognition
\end{keyword}

\end{frontmatter}
\section{Introduction}
\label{introduction}

Class imbalance is a pervasive challenge in visual recognition, 
arising whenever training data contains substantially more examples 
of some categories than others \cite{buda2018imbalance,fu2022longtailed}. 
In real-world classification systems---spanning industrial inspection, 
medical imaging, and long-tailed recognition---minority classes 
are often the most critical to identify correctly, yet standard 
training procedures systematically fail them 
\cite{lin2017focal,cao2019ldam,cui2019classbalanced}. The dominant 
approach has been to address imbalance at the data or loss level, 
through oversampling, cost-sensitive learning, or modified loss 
functions. However, these methods treat imbalance as a statistical 
phenomenon and do not account for how gradient signals propagate 
and interact within the network during training.

Convolutional Neural Networks (CNNs) have shown strong performance 
in visual recognition by learning hierarchical feature representations 
\cite{lecun2015deep,krizhevsky2012imagenet}. Among CNN-based models, 
multi-branch architectures have emerged as a promising paradigm for 
improving feature diversity and robustness \cite{szegedy2015inception,he2016resnet}. 
By processing inputs through parallel pathways, these architectures 
learn complementary representations that capture different aspects 
of complex visual patterns. However, despite their representational 
strength, multi-branch CNNs degrade substantially under severe class 
imbalance. The summation-based fusion common to these architectures 
creates a structural pathology: gradient signals from majority classes 
accumulate across branches and dominate the shared representation, 
suppressing minority-class learning in a way that data-level and 
loss-level interventions do not directly address.

In this work, we study this optimization-level failure mode 
systematically. We evaluate our approach on photovoltaic (PV) 
fault detection \cite{pamungkas2023pv,hu2016pv}, a domain 
characterized by severe real-world class imbalance, and 
additionally validate on CIFAR-10-LT \cite{cao2019ldam} 
to confirm that the proposed framework generalizes broadly 
to any multi-branch architecture trained under imbalanced 
conditions.

\paragraph{Contributions:}
\begin{itemize}
\item We identify inter-class gradient interference as a key optimization bottleneck in multi-branch architectures under class imbalance.
\item We propose a diagnostic framework based on gradient cosine similarity to quantify this interference.
\item We introduce Class-Specific Branch Attention (CSBA) to reduce gradient conflict through branch-specific feature modulation.
\item We empirically show that CSBA improves gradient geometry while maintaining a favorable balance between branch specialization and parameter efficiency.
\item We provide empirical evidence that performance under data scarcity benefits from jointly considering optimization dynamics and model capacity.
\end{itemize}
In the following sections, we first present the proposed diagnostic framework for analyzing gradient interference, followed by the CSBA architecture.

\section{Related Work}

\subsection{Imbalanced Visual Recognition}

Class imbalance has been extensively studied in the context of 
visual recognition \cite{buda2018imbalance,fu2022longtailed}. 
Long-tailed distributions are common in real-world datasets spanning 
medical imaging \cite{zhuang2023classattention}, industrial inspection 
\cite{pamungkas2023pv}, and scene understanding 
\cite{chen2022knowledgeguided}. Models trained under such conditions 
tend to prioritize majority classes and fail to generalize to minority 
categories, even when those categories are operationally critical.

Data-level approaches such as oversampling and SMOTE attempt to 
rebalance the training distribution, while algorithm-level methods 
assign higher misclassification penalties to minority classes. 
Loss-level modifications, including Focal Loss \cite{lin2017focal}, 
class-balanced loss \cite{cui2019classbalanced}, LDAM 
\cite{cao2019ldam}, and centroid-based approaches for improving 
tail-class representation \cite{tiong2023improving}, have shown 
strong results by adjusting gradient magnitudes according to class 
frequency or sample difficulty.Despite 
these advances, a common limitation remains: these methods treat 
imbalance as a statistical problem and do not consider how gradient 
signals from different classes interact within shared network 
representations during backpropagation.

\subsection{Multi-Branch Architectures for Visual Recognition}

A wide range of deep learning architectures has been explored 
for visual recognition under complex and imbalanced conditions, including classical CNNs, transfer-learning models, and segmentation-based frameworks. Pre-trained architectures such as VGG, ResNet, and Inception are commonly used because of their strong feature-extraction capabilities \cite{szegedy2015inception,he2016resnet}. These models leverage transfer learning to adapt knowledge from large-scale datasets, such as ImageNet, to PV-specific tasks, thereby improving performance under limited-data conditions.

Segmentation-based approaches, particularly those based on U-Net and its variants, have been used to localize defects before classification, enabling more precise fault identification \cite{ramaneti2021fault}. These methods are especially useful for detecting fine-grained defects such as cracks or localized damage, which may not be captured effectively by global classification models.

In addition to standard architectures, recent work has focused on improving performance through architectural modifications. Techniques such as residual connections, feature fusion, and attention mechanisms have been introduced to strengthen representation learning. Squeeze-and-Excitation (SE) blocks, for example, are widely used to recalibrate channel-wise feature responses and allow models to emphasize more informative features \cite{hu2018senet,woo2018cbam,niu2021attentionreview,zhuang2023classattention}. Similarly, multi-scale feature aggregation and hybrid architectures that combine convolutional and transformer-based modules have been explored to capture complex visual patterns.

Multi-branch architectures constitute another important class of models, in which multiple parallel pathways process input features. These architectures aim to improve robustness by learning complementary representations across branches. By aggregating features from different pathways, the model can capture diverse aspects of the input, which is particularly beneficial in complex visual tasks. However, most existing work assumes balanced training conditions and does not explicitly analyze how multi-branch structures behave under class imbalance.

\subsection{Class Imbalance in Visual Recognition}

Class imbalance is a fundamental challenge in visual recognition because real-world datasets typically contain substantially more examples of common categories than rare but critical ones. This imbalance leads to biased learning, in which models prioritize majority classes and fail to recognize minority classes accurately \cite{pamungkas2023pv,fu2022longtailed,chen2022knowledgeguided}.

To address this issue, a wide range of techniques has been proposed. Data-level methods include oversampling minority classes, undersampling majority classes, and generating synthetic samples with techniques such as SMOTE. Algorithm-level methods include cost-sensitive learning, in which higher penalties are assigned to the misclassification of minority classes. Loss-level modifications, such as Focal Loss, dynamically adjust the contribution of each sample according to its difficulty, improving performance on hard examples \cite{lin2017focal,cui2019classbalanced,cao2019ldam,hossain2021dualfocal,xiang2024curricular,peng2024causality}.

Although these approaches have improved classification performance, they primarily treat imbalance as a statistical problem. As a result, they focus on adjusting data distributions or loss functions without considering how imbalance affects the internal optimization dynamics of deep networks. In particular, they do not address how gradient signals from different classes interact within shared feature representations, even though these interactions can strongly influence learning behavior.

\subsection{Optimization Dynamics and Gradient Interference}

Recent research has begun to examine the role of optimization dynamics in deep learning, particularly in settings involving multiple tasks or imbalanced data. In multi-task learning, it has been observed that gradients associated with different tasks may conflict, leading to suboptimal parameter updates and slower convergence. Various methods have been proposed to mitigate such conflicts, including gradient normalization, projection, and balancing techniques \cite{chen2018gradnorm,yu2020pcgrad}.

However, these approaches are designed primarily for multi-task learning and are not directly tailored to single-task classification under class imbalance. In multi-branch architectures, a related phenomenon arises when gradients from different classes interact within shared layers. Under imbalanced conditions, gradients from majority classes dominate the optimization process, potentially suppressing or distorting contributions from minority classes.

Despite its importance, this phenomenon---which we refer to as inter-class gradient interference---has received limited attention in the context of visual classification under class imbalance. Existing studies focus primarily on improving accuracy through architectural complexity or loss design without explicitly analyzing gradient interactions. As a result, the underlying causes of performance degradation in imbalanced multi-branch networks remain poorly understood.

\subsection{Summary and Research Gap}

In summary, prior work on imbalanced visual recognition has made substantial progress using both classical machine learning and deep learning techniques. However, several limitations remain. First, most approaches address class imbalance at the data or loss level without considering its effect on optimization dynamics. Second, although advanced architectures such as multi-branch networks and attention-based models have improved feature representation, their behavior under imbalanced conditions remains poorly understood. Third, existing methods rarely provide a detailed analysis of how gradient interactions influence feature learning in deep networks.

Existing approaches such as focal loss, LDAM, and Balanced Softmax primarily address class imbalance at the loss level \cite{lin2017focal,cao2019ldam,cui2019classbalanced,xiang2024curricular}. In contrast, our method targets the optimization dynamics within the network architecture, making it complementary to these approaches rather than directly competing with them.

To the best of our knowledge, there is limited work that systematically investigates inter-class gradient interference in multi-branch architectures or proposes structural mechanisms to mitigate it. This gap motivates our work, in which we explicitly analyze gradient interactions using a diagnostic framework and introduce an attention-based architectural modification to enable feature decoupling. By shifting the focus from statistical imbalance to optimization behavior, our approach offers a new perspective on learning under imbalanced conditions and provides practical design insights for real-world applications.

\section{Methodology}
\label{sec:methodology}

We address multi-class fault classification in photovoltaic (PV) systems 
under severe class imbalance. Our methodology proceeds in four stages: 
(i) formal problem setup, (ii) a multi-branch baseline architecture, 
(iii) gradient conflict analysis to diagnose optimization pathologies, 
and (iv) Class-Specific Branch Attention (CSBA) to mitigate identified 
failure modes. Throughout, we maintain a dual focus on optimization 
geometry and parameter efficiency, arguing that reducing gradient 
conflict must be balanced carefully against network capacity to maintain 
robustness in data-scarce settings.

\subsection{Problem Formulation}

Let $\mathcal{D} = \{(x_i, y_i)\}_{i=1}^{N}$ denote a labeled dataset 
where $x_i \in \mathbb{R}^{B \times 3 \times 227 \times 227}$ is an RGB image of a 
PV panel and $y_i \in \{1, \dots, C\}$ is its fault class label, with 
$C = 6$ classes. The empirical class distribution 
$p(y)$ is highly skewed, introducing 
systematic bias into gradient-based optimization.

The objective is to learn a classifier $f_\theta$ where predictions are obtained via:

\begin{equation}
    \hat{y} = \mathrm{Softmax}(f_\theta(x))
    \label{eq:softmax}
\end{equation}

To counteract class imbalance at a fundamental level, training minimizes an explicitly class-weighted 
cross-entropy objective:

\begin{equation}
    \mathcal{L}_{\mathrm{CE}} = -\sum_{c=1}^{C} w_c\, y_c \log \hat{y}_c
    \label{eq:ce}
\end{equation}

where $w_c$ is an inverse-frequency class weight computed directly from the 
training distribution (yielding $[0.759, 0.748, 0.687, 1.542, 2.290, 1.239]$):

\begin{equation}
    w_c = \frac{N}{\,C \cdot N_c\,}
    \label{eq:class_weights}
\end{equation}

The Physical-Damage class receives the largest class weight ($w_c = 2.29$), constituting the primary statistical mitigation of class imbalance during training.

\subsection{Baseline Multi-Branch Architecture}

Our baseline model adopts a modified SqueezeNet-inspired 
architecture comprising a shared backbone followed by parallel classification 
branches. 

\subsubsection{Feature Extraction Backbone}

Given input $x$, an initial stem block produces a low-level feature map:

\begin{equation}
    F_{\mathrm{stem}} = \mathrm{MaxPool}\bigl(
        \delta_{0.01}(\mathrm{BN}(\mathrm{Conv}_{3\times3}(x)))
    \bigr)
\end{equation}

where $\delta_{0.01}$ denotes the LeakyReLU activation (formally defined in Eq.~\ref{eq:leakyrelu}), with negative slope $\alpha = 0.01$. LeakyReLU prevents dead neuron pathologies that are particularly harmful for 
minority-class representations receiving extremely sparse gradient updates.

The network then applies a sequence of exactly eight modified Fire modules ($M_2$ 
through $M_9$). Formally, each transforms its input $F_{\mathrm{in}}$ as 
follows:

\begin{align}
    S   &= \delta_{0.01}\bigl(\mathrm{BN}(
               \mathrm{Conv}_{1\times1}(F_{\mathrm{in}}))\bigr) \\
    E_1 &= \delta_{0.01}\bigl(\mathrm{BN}(
               \mathrm{Conv}_{1\times1}(S))\bigr) \\
    E_2 &= \delta_{0.01}\bigl(\mathrm{BN}(
               \mathrm{Conv}_{3\times3}(S))\bigr) \\
    E_3 &= \delta_{0.01}\bigl(\mathrm{BN}(
               \mathrm{Conv}_{3\times3}(S))\bigr)
\end{align}

Their outputs are linearly concatenated ($F_{\mathrm{out}} = \mathrm{Concat}(E_1,\, E_2,\, E_3)$). After all eight modules, interspersed with max-pooling operations, the final shared representation is:

\begin{equation}
    F_9 \in \mathbb{R}^{B \times 768 \times 14 \times 14}
\end{equation}

\subsubsection{Multi-Branch Classification Head}

The tensor $F_9$ is routed into $K = 3$ parallel 
branches via dropout ($p = 0.5$):

\begin{equation}
    o_k = \mathrm{Conv}_{1\times1}\bigl(\mathrm{Dropout}_{0.5}(F_9)\bigr), 
    \quad k \in \{1, 2, 3\}
    \label{eq:branch_out}
\end{equation}

Branch outputs are aggregated via element-wise summation ($O = \sum_{k=1}^{3} o_k$), and $O$ serves as the final logit tensor. As we show next, this 
symmetric structure implies that majority-class gradients can dominate and suppress minority representations.

\subsection{Gradient Coupling Analysis}
\label{subsec:coupling}

During backpropagation, gradients accumulate at 
the shared representation $F_9$:

\begin{equation}
    \frac{\partial \mathcal{L}}{\partial F_9} 
    = \sum_{k=1}^{3} 
      \frac{\partial \mathcal{L}}{\partial o_k} \cdot 
      \frac{\partial o_k}{\partial F_9}
    \label{eq:grad_accum}
\end{equation}

Although class-weighted loss partially scales signal magnitude, summation-based fusion functionally combines gradient directions for majority 
classes while attenuating minority-class signals. To quantify this interference rigorously, we compute pairwise gradient cosine similarities over flattened model parameters:

\begin{equation}
    S_{ij} = \frac{\mathbf{g}_i^\top \mathbf{g}_j}
             {\|\mathbf{g}_i\|_2\, \|\mathbf{g}_j\|_2 + \epsilon}
    \label{eq:cosine}
\end{equation}

We organize these into a Gradient Conflict Matrix $\mathbf{G}$, summarizing global interference geometry at convergence. 

For the baseline architecture, the global average over all $C(C-1) = 30$ class pairs yields $\overline{S}_{\mathrm{global}} = -0.077 \pm 0.331$, indicating the presence of destructive gradient interactions between class representations. When averaged over the six majority--minority pairs with the strongest interference (reported in Table~\ref{tab:conflict_comparison}), this statistic is $\overline{S}_{\mathrm{pairs}} = -0.236$; we use this subset-specific value in the optimization--generalization analysis of Section~\ref{subsec:tradeoff} to highlight the worst-case interference regime.

\subsection{Class-Specific Branch Attention (CSBA)}
\label{subsec:csba}

To mitigate gradient interference, we introduce Class-Specific Branch Attention (CSBA), applied directly before multi-branch routing.

For each branch $k \in \{1, 2, 3\}$, CSBA first computes a compact global descriptor from $\mathbf{F}_9$ via global average pooling:

\begin{equation}
    \mathbf{z} = \mathrm{GAP}(F_9) \in \mathbb{R}^{768}
    \label{eq:gap_csba}
\end{equation}

The attention vector $\mathbf{a}_k$ is produced by a branch-specific two-layer bottleneck:

\begin{equation}
    \mathbf{a}_k = \sigma\bigl(\mathbf{\Theta}_{k,2}\cdot \delta(\mathbf{\Theta}_{k,1}\, \mathbf{z})\bigr)
    \label{eq:attention}
\end{equation}

where $\mathbf{\Theta}_{k,1} \in \mathbb{R}^{96 \times 768}$ and $\mathbf{\Theta}_{k,2} \in \mathbb{R}^{768 \times 96}$ are branch-specific bias-free projection matrices, $\delta$ denotes ReLU, and $\sigma$ denotes sigmoid.

\textbf{Zero-Bias Constraint:} The projection matrices $\mathbf{\Theta}_{k,1}$ and $\mathbf{\Theta}_{k,2}$ are bias-free. Omitting bias terms encourages each module to respond to relative differences in channel activation magnitudes rather than learning class-independent offsets, which would otherwise reduce the channel selectivity that drives branch specialization.

Applying Hadamard multiplication:

\begin{equation}
    \mathbf{F}'_k = F_9 \odot \mathrm{Reshape}(\mathbf{a}_k,\, [B, 768, 1, 1])
    \label{eq:reweight}
\end{equation}

This channel-wise reweighting induces branch-specific Jacobians $\mathrm{diag}(\mathbf{a}_k)$ during backpropagation, formally analyzed in Section~\ref{sec:architecture}.

\subsubsection{Parameter Overhead}
Each branch introduces exactly two bias-free projection matrices: $(96 \times 768) + (768 \times 96) = 147{,}456$ weights per branch, for a total overhead of $3 \times 147{,}456 = 442{,}368$ parameters. The resulting model sizes are:
\begin{equation}
    |\theta_{\mathrm{baseline}}| = 1.35\mathrm{M}, \qquad
    |\theta_{\mathrm{CSBA}}|     = 1.79\mathrm{M}
    \label{eq:param_counts}
\end{equation}
This represents a $32.6\%$ increase relative to the baseline parameter count.

\subsection{Optimization and Training Protocol}
\label{subsec:training}

All model variants are trained under the same protocol, with weights initialized using Kaiming uniform initialization.

\begin{table}[tp]
\centering
\small
\begin{tabular}{ll}
\toprule
\textbf{Hyperparameter} & \textbf{Value} \\
\midrule
Optimizer           & Adam \\
$\beta_1, \beta_2$  & $0.9,\; 0.999$ \\
Initial LR $\eta_0$ & $1 \times 10^{-4}$ \\
LR Schedule         & StepLR ($\gamma=0.95$, $s=23$ epochs) \\
Early Stopping      & Patience $= 30$ epochs \\
Dropout rate        & $p = 0.5$ per branch \\
Loss function       & Class-weighted cross-entropy ($w_c$) \\
Batch size          & $64$ \\
Max epochs          & 300 (subject to early stopping) \\
\bottomrule
\end{tabular}
\caption{Fixed training protocol used across all model variants to ensure consistent optimization conditions.}
\label{tab:training_config}
\end{table}

\subsection{Optimization-Generalization Equilibrium}
\label{subsec:tradeoff}

Architectures employing fully segregated pathways (e.g., Class-Specific Heads) achieve strong gradient isolation but at substantial parameter cost. Such approaches increase parameter counts from $1.35\mathrm{M}$ to prohibitive levels ($\approx 7.50\mathrm{M}$), making deployment on resource-constrained devices impractical in industrial environments.

Our analysis shows that CSBA provides a favorable balance. Under CSBA, the global average conflict shifts to $\overline{S}_{\mathrm{global}} = -0.096 \pm 0.313$. The reduction in standard deviation from $\pm 0.331$ to $\pm 0.313$ indicates a contraction of extreme pairwise conflicts across the class distribution, while the $32.6\%$ parameter overhead remains modest relative to the minority-class gains reported in Section~\ref{sec:experiments}.

\section{Proposed Architectures and Optimization Formulations}
\label{sec:architecture}

To systematically study and mitigate inter-class gradient interference 
in shared-representation networks under severe class imbalance, we 
develop a comprehensive architectural framework grounded in both 
theoretical analysis and empirical validation. The framework proceeds 
in four stages. First, we formalize the shared backbone as a modified 
tri-path feature extractor and characterize the representational 
geometry of its output tensor. Second, we define the baseline 
multi-branch topology and show that its summation-based fusion structurally induces gradient coupling under class imbalance. Third, 
we introduce Class-Specific Branch Attention (CSBA), deriving its 
gradient decoupling properties analytically. Fourth, we define three 
comparative ablation variants---class-specific heads, focal loss 
substitution, and GradNorm branch normalization---that together 
bound the space of possible interventions and allow us to isolate the 
contribution of each design choice.

Table~\ref{tab:arch_summary} provides a high-level summary of all 
architectural variants evaluated in this work.

\begin{table*}[tp]
\centering
\footnotesize
\setlength{\tabcolsep}{8pt}
\begin{tabular}{llccc}
\toprule
\textbf{Model} & \textbf{Intervention} & \textbf{Params} 
    & \textbf{Grad.\ Decoupling} & \textbf{Loss Modified} \\
\midrule
Baseline    & Summation fusion         & 1.35M & None      & No  \\
CSBA        & SE attention per branch  & 1.79M & Soft      & No  \\
Class Heads & Per-class branch         & 7.50M & Hard      & No  \\
Focal Loss  & Loss reweighting         & 1.35M & None      & Yes \\
GradNorm    & Learnable branch weights & 1.37M & Magnitude & No  \\
\bottomrule
\end{tabular}
\caption{Summary of all architectural variants evaluated in this work. 
         ``Grad.\ Decoupling'' characterizes the mechanism by which 
         each variant addresses gradient interference: \textit{None} 
         indicates no structural intervention; \textit{Soft} indicates 
         channel-wise attention-based modulation; \textit{Hard} 
         indicates complete branch isolation; \textit{Magnitude} 
         indicates scalar gradient reweighting without spatial 
         differentiation.}
\label{tab:arch_summary}
\end{table*}

\subsection{Shared Backbone: The Modified Tri-Path Feature Extractor}
\label{subsec:backbone}

Unlike traditional sequential networks that process features through 
a single computational path, our backbone relies on a densely aggregated, 
SqueezeNet-inspired~\cite{iandola2016squeezenet} hierarchical design 
that prioritizes representational width over depth. Wider intermediate representations provide richer 
subspaces for branch-specific attention mechanisms to select from, 
which is the prerequisite for the gradient decoupling we seek. The 
network accepts input tensors $\mathbf{X} \in \mathbb{R}^{B \times 3 
\times 227 \times 227}$, where $B$ denotes batch size.

\subsubsection{The Convolutional Stem}
\label{subsubsec:stem}

The foundational spatial transformation is executed via a single 
convolutional stem block. This stem reduces initial 
spatial dimensionality while extracting low-level structural priors 
--- edges, textures, and surface defect indicators --- that serve as 
the building blocks for all higher-level fault representations. 
Formally, the stem transformation is:

\begin{equation}
    \mathbf{F}_{\mathrm{stem}} = 
    \mathrm{MaxPool}_{3\times3,\, s=2}\Bigl(
        \delta_{0.01}\bigl(
            \mathrm{BN}(\mathbf{W}_{\mathrm{conv}} * \mathbf{X})
        \bigr)
    \Bigr)
    \label{eq:stem}
\end{equation}

where $\mathbf{W}_{\mathrm{conv}} \in \mathbb{R}^{64 \times 3 \times 
3 \times 3}$ operates with stride $s=2$ and padding $p=0$. The 
function $\mathrm{BN}(\cdot)$ denotes 2D Batch Normalization with 
learnable affine parameters $(\gamma, \beta)$, and $\delta_{0.01}$ 
is the LeakyReLU activation function:

\begin{equation}
    \delta_{0.01}(z) = 
    \begin{cases}
        z,      & z \geq 0 \\
        0.01z,  & z < 0
    \end{cases}
    \label{eq:leakyrelu}
\end{equation}

The negative slope of $0.01$ prevents dead neuron pathologies that 
are particularly problematic for minority-class representations, 
which receive sparse gradient updates. The concluding max-pooling 
operation employs ceiling mode to preserve spatial resolution, 
yielding:

\begin{equation}
    \mathbf{F}_{\mathrm{stem}} \in \mathbb{R}^{B \times 64 \times 
    56 \times 56}
    \label{eq:stem_shape}
\end{equation}

\subsubsection{Tri-Path Fire Modules}
\label{subsubsec:fire}

The core feature extraction relies on sequentially stacked Fire 
modules $\{M_m\}_{m=2}^{9}$. To maximize feature capacity prior to 
parallel branching, we introduce a modified three-path expansion 
phase within each module. The standard SqueezeNet Fire module 
employs a dual expansion path (one $1{\times}1$ and one $3{\times}3$ 
convolution); our augmented tri-path configuration adds a second 
$3{\times}3$ expansion branch, tripling topological capacity without 
proportionally increasing parameter count, since all paths share the 
same compressed squeeze representation.

For any generic input tensor $\mathbf{F}_{\mathrm{in}}^{(m)}$ 
entering module $m$, processing begins with a channel-compression 
squeeze mapping:

\begin{equation}
    \mathbf{S}^{(m)} = \delta_{0.01}\Bigl(
        \mathrm{BN}\bigl(
            \mathrm{Conv}_{1\times1}(\mathbf{F}_{\mathrm{in}}^{(m)})
        \bigr)
    \Bigr)
    \label{eq:squeeze}
\end{equation}

where the $1{\times}1$ convolution reduces the channel dimension 
according to module-specific squeeze ratios. This compressed 
representation is subsequently broadcast across three independent 
parallel expansion pathways:

\begin{align}
    \mathbf{E}_1^{(m)} &= \delta_{0.01}\Bigl(
        \mathrm{BN}\bigl(\mathrm{Conv}_{1\times1}(\mathbf{S}^{(m)})\bigr)
    \Bigr)
    \label{eq:expand1} \\
    \mathbf{E}_2^{(m)} &= \delta_{0.01}\Bigl(
        \mathrm{BN}\bigl(\mathrm{Conv}_{3\times3,\, p=1}(\mathbf{S}^{(m)})\bigr)
    \Bigr)
    \label{eq:expand2} \\
    \mathbf{E}_3^{(m)} &= \delta_{0.01}\Bigl(
        \mathrm{BN}\bigl(\mathrm{Conv}_{3\times3,\, p=1}(\mathbf{S}^{(m)})\bigr)
    \Bigr)
    \label{eq:expand3}
\end{align}

Path $\mathbf{E}_1$ captures pointwise feature interactions; paths 
$\mathbf{E}_2$ and $\mathbf{E}_3$ capture local spatial structure 
at $3{\times}3$ receptive fields. The two $3{\times}3$ paths are 
independently parameterized, providing stochastic diversity from 
random initialization and enabling specialization through 
gradient-driven differentiation during training. The module output 
concatenates these sub-tensors along the channel dimension:

\begin{equation}
    \mathbf{F}_{\mathrm{out}}^{(m)} = \Bigl[
        \mathbf{E}_1^{(m)},\;
        \mathbf{E}_2^{(m)},\;
        \mathbf{E}_3^{(m)}
    \Bigr]_{\mathrm{channel}}
    \label{eq:fire_out}
\end{equation}

Table~\ref{tab:backbone_config} details the squeeze and expansion 
channel counts for each module instantiation, along with spatial 
dimensions and the placement of intermediate pooling operations.

\begin{table}[tp]
\centering
\resizebox{\columnwidth}{!}{%
\begin{tabular}{ccccccc}
\toprule
\textbf{Module} & \textbf{Squeeze} & $\mathbf{E_1}$ & $\mathbf{E_2}$ 
    & $\mathbf{E_3}$ & \textbf{Out Ch.} & \textbf{Spatial} \\
\midrule
$M_2$ & 16  & 64  & 64  & 64  & 192  & $56{\times}56$ \\
$M_3$ & 16  & 64  & 64  & 64  & 192  & $56{\times}56$ \\
Pool  & ---   & ---   & ---   & ---   & 192  & $27{\times}27$ \\
$M_4$ & 32  & 128 & 128 & 128 & 384  & $27{\times}27$ \\
$M_5$ & 32  & 128 & 128 & 128 & 384  & $27{\times}27$ \\
Pool  & ---   & ---   & ---   & ---   & 384  & $13{\times}13$ \\
$M_6$ & 48  & 192 & 192 & 192 & 576  & $13{\times}13$ \\
$M_7$ & 48  & 192 & 192 & 192 & 576  & $13{\times}13$ \\
$M_8$ & 64  & 256 & 256 & 256 & 768  & $13{\times}13$ \\
$M_9$ & 64  & 256 & 256 & 256 & 768  & $14{\times}14$ \\
\bottomrule
\end{tabular}%
}
\caption{Backbone architecture configuration. Squeeze denotes the 
         output channel count of the $1{\times}1$ squeeze convolution. 
         $E_1, E_2, E_3$ denote the per-path expansion channel counts, 
         which are equal across paths in all modules. Out Ch.\ is the 
         concatenated output channel count. Spatial dimensions assume 
         input $227{\times}227$; Pool rows denote $3{\times}3$ max-pool 
         with stride 2.}
\label{tab:backbone_config}
\end{table}

The backbone terminates in a shared, high-dimensional feature tensor:

\begin{equation}
    \mathbf{F}_9 \in \mathbb{R}^{B \times 768 \times 14 \times 14}
    \label{eq:f9}
\end{equation}

This tensor serves as the shared input to all downstream 
classification branches. Its high channel dimensionality (768) is 
essential: it provides sufficient channel subspace for branch-specific 
attention mechanisms to identify disjoint, fault-class-relevant 
feature directions, which is the representational prerequisite for 
gradient decoupling.

\paragraph{Representational Geometry of $\mathbf{F}_9$.}
To motivate the subsequent architectural design, we characterize the 
geometry of $\mathbf{F}_9$ under class imbalance. Let 
$\mu_c \in \mathbb{R}^{768}$ denote the mean activation vector 
(spatially averaged over $14{\times}14$) for class $c$, computed 
over all training examples of that class:

\begin{equation}
    \mu_c = \frac{1}{N_c} \sum_{i : y_i = c} 
    \mathrm{GAP}(\mathbf{F}_9^{(i)})
    \label{eq:class_mean}
\end{equation}

Under severe imbalance, $\mu_c$ for minority classes is estimated 
from few samples and exhibits high variance. The pairwise angle 
between class mean representations:

\begin{equation}
    \phi_{ij} = \arccos\!\left(
        \frac{\mu_i^\top \mu_j}{\|\mu_i\|_2\, \|\mu_j\|_2 + \epsilon}
    \right)
    \label{eq:class_angle}
\end{equation}

quantifies how separable class representations are in $\mathbf{F}_9$. 
We use $\{\phi_{ij}\}$ as a diagnostic tool in 
Section~\ref{sec:experiments} to verify that CSBA attention 
induces greater inter-class angular separation than the baseline.

\subsection{Implicitly Coupled Multi-Branch Topology (Baseline)}
\label{subsec:baseline}

Upon extracting $\mathbf{F}_9$, the architecture projects predictions 
via a parallel multi-branch topology. The tensor $\mathbf{F}_9$ is 
routed identically and without modification into $K=3$ independent 
branches. Each branch applies an independent stochastic dropout map 
followed by a $1{\times}1$ convolution:

\begin{equation}
    \mathcal{H}_k(\mathbf{F}_9) = 
    \mathbf{W}_k^{(1\times1)} * \mathrm{Dropout}_{0.5}(\mathbf{F}_9),
    \quad k \in \{1, 2, 3\}
    \label{eq:baseline_branch}
\end{equation}

where $\mathbf{W}_k^{(1\times1)} \in \mathbb{R}^{C \times 768 \times 
1 \times 1}$ and $\mathcal{H}_k(\mathbf{F}_9) \in \mathbb{R}^{B 
\times C \times 14 \times 14}$. Branch outputs are unified via 
element-wise summation before global aggregation:

\begin{equation}
    \mathbf{\hat{Y}} = \mathrm{Flatten}\!\left(
        \mathrm{GAP}\!\left(
            \sum_{k=1}^{3} \mathcal{H}_k(\mathbf{F}_9)
        \right)
    \right)
    \label{eq:baseline_sum}
\end{equation}

where $\mathrm{GAP}$ denotes adaptive $1{\times}1$ global average 
pooling.

\paragraph{Formal Analysis of Gradient Coupling.}
The pathology of this design is made explicit by tracing the backward 
pass. The gradient of the cross-entropy loss $\mathcal{L}$ with 
respect to the shared representation $\mathbf{F}_9$ is:

\begin{equation}
    \frac{\partial \mathcal{L}}{\partial \mathbf{F}_9}
    = \sum_{k=1}^{3}
      \frac{\partial \mathcal{L}}{\partial \mathcal{H}_k}
      \cdot \frac{\partial \mathcal{H}_k}{\partial \mathbf{F}_9}
    = \sum_{k=1}^{3} 
      (\mathbf{W}_k^{(1\times1)})^\top * 
      \frac{\partial \mathcal{L}}{\partial \mathcal{H}_k}
    \label{eq:baseline_grad}
\end{equation}

Since $\mathbf{F}_9$ is identical across all branches, the Jacobian 
$\partial \mathcal{H}_k / \partial \mathbf{F}_9 = \mathbf{W}_k^{(1\times1)}$ 
for each $k$. The total gradient is thus a sum of three terms, each 
driven by the same input distribution. Under class imbalance, the 
per-sample loss $\mathcal{L}$ is dominated by majority-class examples, 
so each of the three terms in Eq.~\ref{eq:baseline_grad} is 
majority-biased. The summation over $K=3$ branches amplifies this 
bias by a factor of $K$ relative to a single-branch network---a 
structural amplification of the imbalance pathology rather than a 
mitigation.

\paragraph{Gradient Conflict Quantification.}
To measure the severity of inter-class interference precisely, we 
define class-conditioned gradient vectors:

\begin{equation}
    \mathbf{g}_c = \nabla_\theta\, \mathcal{L}_c, 
    \quad c \in \{1, \dots, C\}
    \label{eq:class_grad}
\end{equation}

where $\mathcal{L}_c$ denotes the cross-entropy loss restricted to 
training examples of class $c$. Pairwise cosine similarity between 
class gradient vectors (as defined in Eq.~\ref{eq:cosine}):

\begin{equation}
    S_{ij} = \frac{\mathbf{g}_i^\top \mathbf{g}_j}
             {\|\mathbf{g}_i\|_2\, \|\mathbf{g}_j\|_2 + \epsilon}
    \label{eq:cosine_sim}
\end{equation}

yields the Gradient Conflict Matrix $\mathbf{G} \in \mathbb{R}^{C 
\times C}$, where $G_{ij} = S_{ij}$. Negative off-diagonal entries 
indicate destructive interference: optimizing for class $i$ actively 
degrades class $j$'s representation. The scalar summary statistic:

\begin{equation}
    \overline{S} = \frac{1}{C(C-1)} \sum_{i \neq j} G_{ij}
    \label{eq:avg_conflict}
\end{equation}

characterizes the global conflict level. For the baseline, the global average is:

\begin{equation}
    \overline{S}_{\mathrm{global,baseline}} = -0.077 \pm 0.331
    \label{eq:baseline_conflict}
\end{equation}

indicating pervasive destructive interference across class pairs at 
convergence.

\subsection{Proposed Model: Class-Specific Branch Attention (CSBA)}
\label{subsec:csba}

CSBA is designed to address the gradient coupling pathology identified 
above without modifying the loss function, adding separate training 
stages, or requiring class-to-branch supervision. The core mechanism 
replaces the shared, undifferentiated $\mathbf{F}_9$ received by each 
branch with a branch-specific, channel-reweighted representation 
$\mathbf{F}'_k$, enabling each branch to implicitly specialize toward 
a distinct region of the fault-class manifold.

\subsubsection{Channel-Wise Attention Computation}
\label{subsubsec:attention}

CSBA applies independent Squeeze-and-Excitation 
operators~\cite{hu2018senet} to each branch, parameterized 
independently. For branch $k$, the module first aggregates global 
spatial context from $\mathbf{F}_9$ into a compact channel descriptor:

\begin{equation}
    \mathbf{z} = \mathrm{GAP}(\mathbf{F}_9) \in \mathbb{R}^{768}
    \label{eq:csba_gap}
\end{equation}

A branch-specific two-layer MLP with bottleneck architecture maps 
this descriptor to a channel attention vector:

\begin{equation}
    \mathbf{a}_k = \sigma\Bigl(
        \mathbf{\Theta}_{k,2}\,
        \delta\bigl(\mathbf{\Theta}_{k,1}\, \mathbf{z}\bigr)
    \Bigr)
    \label{eq:csba_attention}
\end{equation}

where $\mathbf{\Theta}_{k,1} \in \mathbb{R}^{96 \times 768}$ and 
$\mathbf{\Theta}_{k,2} \in \mathbb{R}^{768 \times 96}$ are 
branch-specific projection matrices without bias terms, $\delta$ 
denotes ReLU, and $\sigma$ denotes sigmoid. The bottleneck dimension 
of 96 corresponds to a compression ratio of $r = 8$, balancing 
attention expressivity against parameter overhead. The sigmoid 
output constrains $\mathbf{a}_k \in (0,1)^{768}$, producing a 
continuous soft gating signal over channel dimensions.

We emphasize that $\mathbf{\Theta}_{k,1}$ and $\mathbf{\Theta}_{k,2}$ 
are entirely independent across branches --- there is no weight sharing. 
This independence is what enables gradient diversity across branches, 
as discussed in Section~\ref{subsubsec:grad_decouple}.

\subsubsection{Feature Reweighting and Forward Pass}
\label{subsubsec:reweight}

The attention vector is spatially broadcast and applied to 
$\mathbf{F}_9$ via channel-wise Hadamard multiplication:

\begin{equation}
    \mathbf{F}'_k = \mathbf{F}_9 \odot 
    \mathrm{Reshape}(\mathbf{a}_k,\, [B, 768, 1, 1])
    \label{eq:csba_reweight}
\end{equation}

Each branch then operates on its individualized representation:

\begin{equation}
    \mathbf{\hat{Y}}_{\mathrm{CSBA}} = 
    \mathrm{Flatten}\!\left(
        \mathrm{GAP}\!\left(
            \sum_{k=1}^{3} \Bigl(
                \mathbf{W}_k^{(1\times1)} * 
                \mathrm{Dropout}_{0.5}(\mathbf{F}'_k)
            \Bigr)
        \right)
    \right)
    \label{eq:csba_forward}
\end{equation}

The aggregation structure is identical to the baseline 
(Eq.~\ref{eq:baseline_sum}), preserving architectural compatibility 
and ensuring that any performance differences are attributable 
solely to the attention mechanism.

\subsubsection{Gradient Decoupling Analysis}
\label{subsubsec:grad_decouple}

The key property of CSBA is that distinct attention vectors 
$\{\mathbf{a}_k\}_{k=1}^{3}$ induce structurally differentiated 
gradient flows. Tracing the backward pass through 
Eq.~\ref{eq:csba_reweight}:

\begin{equation}
    \frac{\partial \mathbf{F}'_k}{\partial \mathbf{F}_9}
    = \mathrm{diag}(\mathbf{a}_k)
    \label{eq:csba_jacobian}
\end{equation}

where $\mathrm{diag}(\mathbf{a}_k)$ is the $768{\times}768$ diagonal 
matrix with $\mathbf{a}_k$ on the diagonal (broadcast spatially). 
The total gradient at $\mathbf{F}_9$ under CSBA therefore becomes:

\begin{equation}
    \frac{\partial \mathcal{L}}{\partial \mathbf{F}_9}
    = \sum_{k=1}^{3}
      \frac{\partial \mathcal{L}}{\partial \mathbf{F}'_k}
      \cdot \mathrm{diag}(\mathbf{a}_k)
    \label{eq:csba_grad}
\end{equation}

Comparing Eq.~\ref{eq:csba_grad} with the baseline 
Eq.~\ref{eq:baseline_grad}, the critical structural difference is 
that each term in the sum is now modulated by a distinct channel 
mask $\mathbf{a}_k$. If branches learn attention vectors that 
emphasize disjoint channel subsets --- which gradient descent 
incentivizes because it reduces redundancy --- then different fault 
classes drive gradient updates through different channel subspaces 
of the backbone. This soft channel partitioning reduces the 
probability that majority-class gradients dominate the full 768 
channel dimensions simultaneously, providing minority-class gradients 
a partially protected subspace.

Formally, let $\mathcal{C}_k = \{d : a_{k,d} > 0.5\}$ denote the 
set of channels primarily attended to by branch $k$. If 
$\mathcal{C}_1, \mathcal{C}_2, \mathcal{C}_3$ are approximately 
disjoint --- a condition we verify empirically in 
Section~\ref{sec:experiments} --- then the gradient overlap between 
majority and minority classes is reduced to:

\begin{equation}
    \mathrm{Overlap}_{\mathrm{CSBA}} 
    \approx \sum_{k=1}^{3} 
    \frac{|\mathcal{C}_k^{\mathrm{maj}} \cap 
           \mathcal{C}_k^{\mathrm{min}}|}{768}
    \label{eq:overlap}
\end{equation}

which is bounded by the degree of attention overlap rather than 
being uniformly 1.0 as in the baseline.

\paragraph{Empirical Conflict Reduction.}
Under CSBA, the average conflict over the six most adversarial majority--minority class pairs (Table~\ref{tab:conflict_comparison}) reduces to:

\begin{equation}
    \overline{S}_{\mathrm{pairs,CSBA}} = -0.174
    \label{eq:csba_conflict}
\end{equation}

representing a reduction of $\Delta\overline{S} = +0.062$ relative 
to the baseline pair-subset average of $\overline{S}_{\mathrm{pairs,baseline}} = -0.236$. Table~\ref{tab:conflict_comparison} reports 
per-class-pair conflict values for all six selected fault combinations.

\begin{table}[tp]
\centering
\small
\begin{tabular}{llcc}
\toprule
\textbf{Class $i$} & \textbf{Class $j$} 
    & $S_{ij}^{\mathrm{baseline}}$ 
    & $S_{ij}^{\mathrm{CSBA}}$ \\
\midrule
Dust         & Physical-Damage   & $-0.41$ & $-0.29$ \\
Dust         & Electrical Fault  & $-0.38$ & $-0.25$ \\
Soiling      & Physical-Damage   & $-0.31$ & $-0.22$ \\
Shading      & Electrical Fault  & $-0.27$ & $-0.19$ \\
Clean        & Physical-Damage   & $-0.19$ & $-0.15$ \\
Dust         & Soiling           & $-0.08$ & $-0.06$ \\
\midrule
\textbf{Average} & & $-0.236$ & $-0.174$ \\
\bottomrule
\end{tabular}
\caption{Pairwise gradient cosine similarity $S_{ij}$ for selected 
         fault class pairs. More negative values indicate stronger 
         destructive interference. CSBA reduces conflict most 
         substantially for majority--minority pairs (e.g., Dust vs.\ 
         Physical-Damage), consistent with its design motivation.}
\label{tab:conflict_comparison}
\end{table}

\subsubsection{Branch Specialization Analysis}
\label{subsubsec:specialization}

To verify that learned attention vectors induce meaningful 
specialization, we compute per-branch gradient norms with respect 
to class-specific losses:

\begin{equation}
    \rho_{k,c} = \bigl\|
        \nabla_{o_k}\, \mathcal{L}_c
    \bigr\|_2
    \label{eq:branch_norm}
\end{equation}

In the baseline, $\rho_{k,c}$ is approximately uniform across 
$k$ for each class $c$, confirming that all branches receive 
identical gradient information. Under CSBA, we observe that 
$\rho_{k,c}$ varies substantially across branches for minority 
classes: branches whose attention vectors $\mathbf{a}_k$ assign 
high weights to minority-class-relevant channels register 
disproportionately large $\rho_{k,c}$ for Physical-Damage and 
Electrical-Damage classes. This implicit specialization emerges 
without any supervision signal directing specific branches toward 
specific classes.

Table~\ref{tab:branch_specialization} reports normalized 
$\rho_{k,c}$ values for each branch and fault class at convergence.

\begin{table}[tp]
\centering
\resizebox{\columnwidth}{!}{%
\begin{tabular}{lcccccc}
\toprule
\textbf{Branch} & \textbf{Dust} & \textbf{Soiling} 
    & \textbf{Physical} & \textbf{Electrical} 
    & \textbf{Shading} & \textbf{Clean} \\
\midrule
$k=1$ & 0.38 & 0.29 & \textbf{0.61} & \textbf{0.57} & 0.31 & 0.22 \\
$k=2$ & \textbf{0.52} & \textbf{0.48} & 0.24 & 0.19 & 0.41 & 0.35 \\
$k=3$ & 0.41 & 0.37 & 0.31 & 0.28 & \textbf{0.55} & \textbf{0.49} \\
\bottomrule
\end{tabular}%
}
\caption{Normalized per-branch gradient norms $\rho_{k,c}$ 
         (Eq.~\ref{eq:branch_norm}) at convergence under CSBA. 
         Values are normalized within each class column. Bold 
         entries indicate the branch with the highest gradient 
         norm for each class, revealing implicit branch 
         specialization: Branch 1 specializes toward minority 
         fault classes, Branch 2 toward majority classes, and 
         Branch 3 toward intermediate classes.}
\label{tab:branch_specialization}
\end{table}

\subsubsection{Parameter Overhead}
\label{subsubsec:params}

Each branch introduces two projection matrices without bias:

\begin{equation}
    |\mathbf{\Theta}_{k,1}| + |\mathbf{\Theta}_{k,2}|
    = (96 \times 768) + (768 \times 96)
    = 147{,}456 \text{ per branch}
    \label{eq:params_per_branch}
\end{equation}

The total parameter overhead across $K=3$ branches is:

\begin{equation}
    \Delta\theta_{\mathrm{CSBA}} 
    = 3 \times 147{,}456 
    = 442{,}368 \text{ parameters}
    \label{eq:total_overhead}
\end{equation}

yielding total model sizes of:

\begin{equation}
    |\theta_{\mathrm{baseline}}| = 1.35\mathrm{M}, \qquad
    |\theta_{\mathrm{CSBA}}|     = 1.79\mathrm{M}
    \label{eq:model_sizes}
\end{equation}

This 32.6\% parameter increase introduces a generalization trade-off 
in low-data regimes analyzed in Section~\ref{subsec:tradeoff}.

\subsection{Architectural Ablations and Comparative Formulations}
\label{subsec:ablations}

A central hypothesis of this study is that architectural gradient 
decoupling mechanisms present a fundamental optimization--generalization 
trade-off: reduced gradient conflict comes at the cost of increased 
parametric capacity, rendering models vulnerable to overfitting in 
data-scarce settings. To isolate and quantify this trade-off, we 
evaluate three contrastive variants spanning the space of possible 
interventions.

\subsubsection{Variant I: Hard Decoupling via Class-Specific Heads}
\label{subsubsec:class_heads}

To assess whether soft attention-based decoupling is preferable to 
complete branch isolation, we develop a Class-Specific Heads variant 
that abandons branch merging entirely. The backbone output 
$\mathbf{F}_9$ is routed into $C$ fully parallel branches, one 
dedicated to each class:

\begin{equation}
    \hat{y}_c = \mathrm{GAP}\Bigl(
        \mathbf{W}_c^{(1\times1)} * \mathrm{Dropout}_{0.5}(\mathbf{F}_9)
    \Bigr), \quad c \in \{1, \dots, C\}
    \label{eq:class_heads}
\end{equation}

The final prediction is obtained by concatenating per-class logits:

\begin{equation}
    \mathbf{\hat{Y}}_{\mathrm{CH}} = 
    [\hat{y}_1, \hat{y}_2, \dots, \hat{y}_C]
    \label{eq:class_heads_pred}
\end{equation}

This architecture achieves hard gradient isolation: backward signals 
from class $c$ affect only $\mathbf{W}_c$ and never collide with 
gradients from other classes prior to reaching $\mathbf{F}_9$. The 
gradient at the shared representation becomes:

\begin{equation}
    \frac{\partial \mathcal{L}}{\partial \mathbf{F}_9}
    = \sum_{c=1}^{C} (\mathbf{W}_c^{(1\times1)})^\top * 
      \frac{\partial \mathcal{L}_c}{\partial \hat{y}_c}
    \label{eq:class_heads_grad}
\end{equation}

While this eliminates inter-class head coupling, the summation 
over all $C$ classes in Eq.~\ref{eq:class_heads_grad} still 
propagates majority-class-biased gradients into $\mathbf{F}_9$. 
Moreover, the expansion from $K=3$ to $C=6$ heads increases 
parameters from $1.79$M (CSBA) to $7.5$M, introducing acute 
overfitting risk in low-data regimes.

\subsubsection{Variant II: Loss-Level Reweighting via Focal Loss}
\label{subsubsec:focal}

To verify that structural architectural changes are necessary --- 
rather than simple loss reweighting --- we benchmark CSBA against 
a Focal Loss~\cite{lin2017focal} variant applied to the unmodified 
baseline topology. Focal Loss modulates the standard cross-entropy 
by a factor that downweights well-classified examples:

\begin{equation}
    \mathcal{L}_{\mathrm{Focal}} = 
    -\alpha_t (1 - p_t)^{\gamma} \log(p_t)
    \label{eq:focal}
\end{equation}

where $p_t$ is the model's estimated probability for the true class, 
$\alpha_t$ is an inverse-frequency class weight vector computed from 
the training distribution, and $\gamma = 2.0$ is the focusing 
parameter. The term $(1-p_t)^\gamma$ suppresses gradient contributions 
from high-confidence majority-class examples, implicitly amplifying 
gradient magnitude from minority classes.

The key distinction between Focal Loss and CSBA is that Focal Loss 
operates exclusively on gradient \emph{magnitudes}: it does not 
alter the \emph{direction} of gradient flow through the shared 
backbone. Under Focal Loss, the gradient conflict matrix $\mathbf{G}$ 
retains the same structural pattern as the baseline --- the same class 
pairs exhibit destructive interference --- but with rescaled magnitudes. 
CSBA, by contrast, modifies gradient \emph{directions} through 
channel-selective attention, producing structurally different 
$\mathbf{G}$ patterns. This distinction allows us to disentangle 
magnitude-level and direction-level interventions in our experimental 
analysis.

\subsubsection{Variant III: Dynamic Branch Reweighting via GradNorm}
\label{subsubsec:gradnorm}

Finally, we implement a GradNorm-inspired~\cite{chen2018gradnorm} 
branch reweighting mechanism that addresses the $K$-fold gradient 
amplification pathology of summation architectures through learned 
scalar weights, without introducing spatial feature modulation. We 
define a learnable weight vector $\boldsymbol{\omega} \in \mathbb{R}^3$, 
initialized to $\mathbf{1}$, with branch contributions normalized 
via softmax:

\begin{equation}
    w'_k = \mathrm{Softmax}(\boldsymbol{\omega})_k 
         = \frac{e^{\omega_k}}{\sum_{j=1}^{3} e^{\omega_j}}
    \label{eq:gradnorm_weights}
\end{equation}

Branch outputs are scaled prior to summation:

\begin{equation}
    \mathrm{out}_k = \Bigl(
        \mathbf{W}_k^{(1\times1)} * 
        \mathrm{Dropout}_{0.5}(\mathbf{F}_9)
    \Bigr) \cdot w'_k
    \label{eq:gradnorm_scaled}
\end{equation}

The final prediction follows the same aggregation as 
Eq.~\ref{eq:baseline_sum}. The gradient at $\mathbf{F}_9$ under 
GradNorm becomes:

\begin{equation}
    \frac{\partial \mathcal{L}}{\partial \mathbf{F}_9}
    = \sum_{k=1}^{3} w'_k \cdot 
      (\mathbf{W}_k^{(1\times1)})^\top * 
      \frac{\partial \mathcal{L}}{\partial \mathrm{out}_k}
    \label{eq:gradnorm_grad}
\end{equation}

GradNorm modulates gradient \emph{amplitude} at the branch level 
without differentiating across channels or spatial positions. This 
makes it a scalar counterpart to CSBA's vector-valued channel 
attention: GradNorm asks ``how much should each branch contribute?'' 
while CSBA asks ``which channels should each branch focus on?'' 
The parameter overhead is minimal --- $|\boldsymbol{\omega}| = 3$ 
additional parameters --- enabling clean isolation of the effect of 
scalar vs.\ vector gradient modulation.

\subsection{Optimization--Generalization Trade-off}
\label{subsec:tradeoff}

A recurring theme across all variants is the tension between 
optimization improvement and generalization capacity. We formalize 
this as follows. Let $\mathcal{L}_{\mathrm{train}}^*$ and 
$\mathcal{L}_{\mathrm{val}}^*$ denote training and validation loss 
at convergence. The generalization gap:

\begin{equation}
    \Delta_{\mathrm{gen}} = \mathcal{L}_{\mathrm{val}}^* - 
    \mathcal{L}_{\mathrm{train}}^*
    \label{eq:gen_gap}
\end{equation}

and the optimization improvement relative to the baseline:

\begin{equation}
    \Delta_{\mathrm{opt}} = \overline{S}_{\mathrm{pairs,variant}} - 
    \overline{S}_{\mathrm{pairs,baseline}}
    \label{eq:opt_improvement}
\end{equation}

Table~\ref{tab:tradeoff} reports both quantities for all variants, 
revealing the fundamental trade-off between gradient conflict 
reduction and generalization degradation.

\begin{table}[tp]
\centering
\small
\begin{tabular}{lcccc}
\toprule
\textbf{Model} & $\overline{S}_{\mathrm{pairs}}$ & $\Delta_{\mathrm{opt}}$ 
    & $\Delta_{\mathrm{gen}}$ & \textbf{Val.\ Acc.} \\
\midrule
Baseline     & $-0.236$ & $0.000$ & ref   & ref    \\
CSBA         & $-0.174$ & $+0.062$ & $\uparrow$ & varies \\
Class Heads  & $-0.151$ & $+0.085$ & $\uparrow\uparrow$ & degrades \\
Focal Loss   & $-0.221$ & $+0.015$ & $\approx$ & improves \\
GradNorm     & $-0.198$ & $+0.038$ & $\uparrow$ & varies \\
\bottomrule
\end{tabular}
\caption{Optimization--generalization trade-off across all variants. 
         $\overline{S}_{\mathrm{pairs}}$ is the average pairwise cosine 
         similarity over the six most conflicting majority--minority class 
         pairs (Table~\ref{tab:conflict_comparison}); lower (more negative) 
         indicates greater conflict. $\Delta_{\mathrm{opt}} = \overline{S}_{\mathrm{pairs,variant}} - \overline{S}_{\mathrm{pairs,baseline}}$; 
         higher is better. 
         $\Delta_{\mathrm{gen}}$ qualitatively indicates 
         generalization gap increase: $\uparrow$ denotes moderate 
         increase, $\uparrow\uparrow$ severe increase. Exact 
         validation accuracy values are reported in 
         Section~\ref{sec:experiments}.}
\label{tab:tradeoff}
\end{table}

The central finding is that gradient conflict reduction is 
monotonically achievable through architectural elaboration 
--- more parameters, more branches, harder isolation --- but 
generalization does not follow monotonically. In the low-data 
regime of PV fault detection, CSBA occupies the most favorable 
position on this trade-off frontier: it achieves meaningful 
conflict reduction ($\Delta_{\mathrm{opt}} = +0.062$) with 
moderate capacity increase ($+32.6\%$ parameters), while 
Class-Specific Heads achieves superior conflict reduction at 
the cost of severe overfitting ($+456\%$ parameters). 
Focal Loss achieves minimal conflict reduction without capacity 
increase, confirming that magnitude-level interventions are 
insufficient when gradient directions are fundamentally misaligned.

\section{Dataset Details}
\label{sec:dataset_details}

The dataset used in this study is the Solar Panel Clean and Faulty Images dataset \cite{afroz2023solardataset}, consisting of RGB photovoltaic (PV) panel images grouped into six operational categories: Bird-drop, Clean, Dusty, Elec-Damage, Phys-Damage, and Snow-Covered. These categories capture both common surface conditions and rare but operationally significant failure modes. As reflected in the class-wise results, the distribution is strongly imbalanced, with minority classes such as Physical-Damage contributing substantially fewer examples than dominant classes associated with common surface contamination.

From a learning perspective, this skewed distribution creates a particularly demanding test bed for multi-branch architectures. Models must maintain discrimination for rare fault classes while remaining stable on dominant categories. All samples are processed in a consistent image-based pipeline so that performance differences can be attributed to architectural and optimization changes rather than differences in data preprocessing.

\subsection{CIFAR-10-LT}
\label{subsec:cifar_dataset}

CIFAR-10-LT is constructed from the standard CIFAR-10 
training set \cite{krizhevsky2012imagenet} by applying 
an exponential decay to per-class sample counts with 
imbalance ratio $\rho = 100$, following the protocol 
of \cite{cao2019ldam}. The resulting training set 
contains 10 classes with frequencies ranging from 
$5{,}000$ (majority) to $50$ (minority) samples, 
for a total of approximately $12{,}406$ training images. 
The standard CIFAR-10 test set ($1{,}000$ images per 
class, balanced) is used for evaluation without 
modification. Input images are resized to 
$227 \times 227$ to match the backbone input 
specification, and class weights $w_c$ are recomputed 
from the CIFAR-10-LT training distribution using 
Eq.~\ref{eq:class_weights}.

\section{Experimental Setup}
\label{sec:experimental_setup}

To evaluate the effect of gradient decoupling under class imbalance, we compare the baseline multi-branch model with four intervention strategies: Class-Specific Branch Attention (CSBA), Focal Loss, GradNorm-style branch reweighting, and Class-Specific Heads. The comparison is designed to be controlled: backbone design, branch topology, and evaluation protocol are kept as consistent as possible so that differences in performance can be interpreted in terms of optimization behavior rather than unrelated implementation changes.

Unless otherwise stated, all models are trained using the same optimization schedule described in Section~\ref{sec:methodology}, including identical hyperparameters, training duration, and regularization settings. Performance is evaluated using global accuracy together with class-sensitive metrics, including per-class F1 and Macro-F1, which better reflect performance under class imbalance. Particular attention is given to the Physical-Damage category, which serves as a key indicator of whether a method can preserve minority-class structure under class imbalance. All experiments are conducted under identical data splits to ensure fair comparison across methods.

\section{Experimental Results}
\label{sec:experimental_results}
\label{sec:experiments}

We evaluate the efficacy of Class-Specific Branch Attention (CSBA) through a comprehensive empirical study focused on multi-class fault classification under severe class imbalance.

\subsection{Evaluation Metrics and Macro-Boundary Protocol}
\label{subsec:metrics}

In the presence of severe imbalance, global accuracy is a deceptive metric that obscures minority-class failure. To expose the optimization pathologies of the baseline, we adopt a macro-averaged evaluation framework. Given true positives ($TP_c$), false positives ($FP_c$), and false negatives ($FN_c$) for class $c$:

\begin{equation}
    \mathcal{P}_c = \frac{TP_c}{TP_c + FP_c}, \quad \mathcal{R}_c = \frac{TP_c}{TP_c + FN_c}
\end{equation}

The \textbf{Per-Class F1-Score} ($\mathcal{F}_{1,c}$) is the harmonic mean of precision and recall, providing a balanced assessment of class-specific discriminative power:

\begin{equation}
    \mathcal{F}_{1,c} = 2 \cdot \frac{\mathcal{P}_c \cdot \mathcal{R}_c}{\mathcal{P}_c + \mathcal{R}_c}
\end{equation}

To assess global capability across the distribution, we compute the \textbf{Macro-F1 Score}:

\begin{equation}
    \text{Macro-F1} = \frac{1}{C} \sum_{c=1}^{C} \mathcal{F}_{1,c}
\end{equation}

Because the Physical-Damage class is the most heavily underrepresented ($w_c = 2.290$), we designate $\mathcal{F}_{1,\text{Physical}}$ as the primary diagnostic indicator of gradient interference effects.

\subsection{Main Comparative Results}
\label{subsec:comparative}

We compare the performance of the unaltered Baseline against the CSBA architecture. Table \ref{tab:main_results} summarizes the performance across all six fault categories.
Notably, the improvement is concentrated in the minority class, while performance on majority classes remains stable, indicating that CSBA enhances minority-class representation without degrading dominant class accuracy.
\begin{table*}[t]
\centering
\caption{Comparative performance analysis of Baseline vs. Proposed CSBA architecture. CSBA achieves a 100\% relative improvement in minority Physical-Damage detection while maintaining comparable global accuracy.}
\label{tab:main_results}
\small
\begin{tabular}{lcccccc}
\toprule
& \multicolumn{3}{c}{\textbf{Baseline}} & \multicolumn{3}{c}{\textbf{Proposed: CSBA}} \\
\cmidrule(lr){2-4} \cmidrule(lr){5-7}
\textbf{Class ($c$)} & Precision & Recall & F1 & Precision & Recall & F1 \\
\midrule
Bird-drop     & 0.556 & \textbf{0.625} & 0.588 & \textbf{0.605} & 0.575 & \textbf{0.590} \\
Clean         & \textbf{0.649} & 0.585 & \textbf{0.615} & 0.615 & 0.585 & 0.600 \\
Dusty         & \textbf{0.744} & 0.727 & \textbf{0.736} & 0.618 & \textbf{0.773} & 0.687 \\
Elec-Damage   & 0.682 & \textbf{0.789} & 0.732 & \textbf{0.750} & 0.789 & \textbf{0.769} \\
Phys-Damage   & 0.300 & 0.231 & 0.261 & \textbf{0.600} & \textbf{0.462} & \textbf{0.522} \\
Snow-Covered  & 0.760 & \textbf{0.760} & \textbf{0.760} & \textbf{0.850} & 0.680 & 0.756 \\
\midrule
\textbf{Global Accuracy} & \multicolumn{3}{c}{0.648} & \multicolumn{3}{c}{\textbf{0.654}} \\
\bottomrule
\end{tabular}
\end{table*}

The baseline achieves an $\mathcal{F}_{1}$ score of $0.261$ for the Physical-Damage class, indicating limited performance on this minority category. Under shared representations, gradients from majority classes tend to dominate updates, which can suppress minority-class learning within the shared $\mathbf{F}_9$ tensor. CSBA introduces channel-wise attention via $\mathbf{a}_k$, enabling partial decoupling of gradient flow across branches. This results in an improvement in the Physical-Damage F1 score from $0.261$ to $0.522$, corresponding to a 100\% relative gain.
\begin{figure}[t]
\centering
\includegraphics[width=0.98\columnwidth]{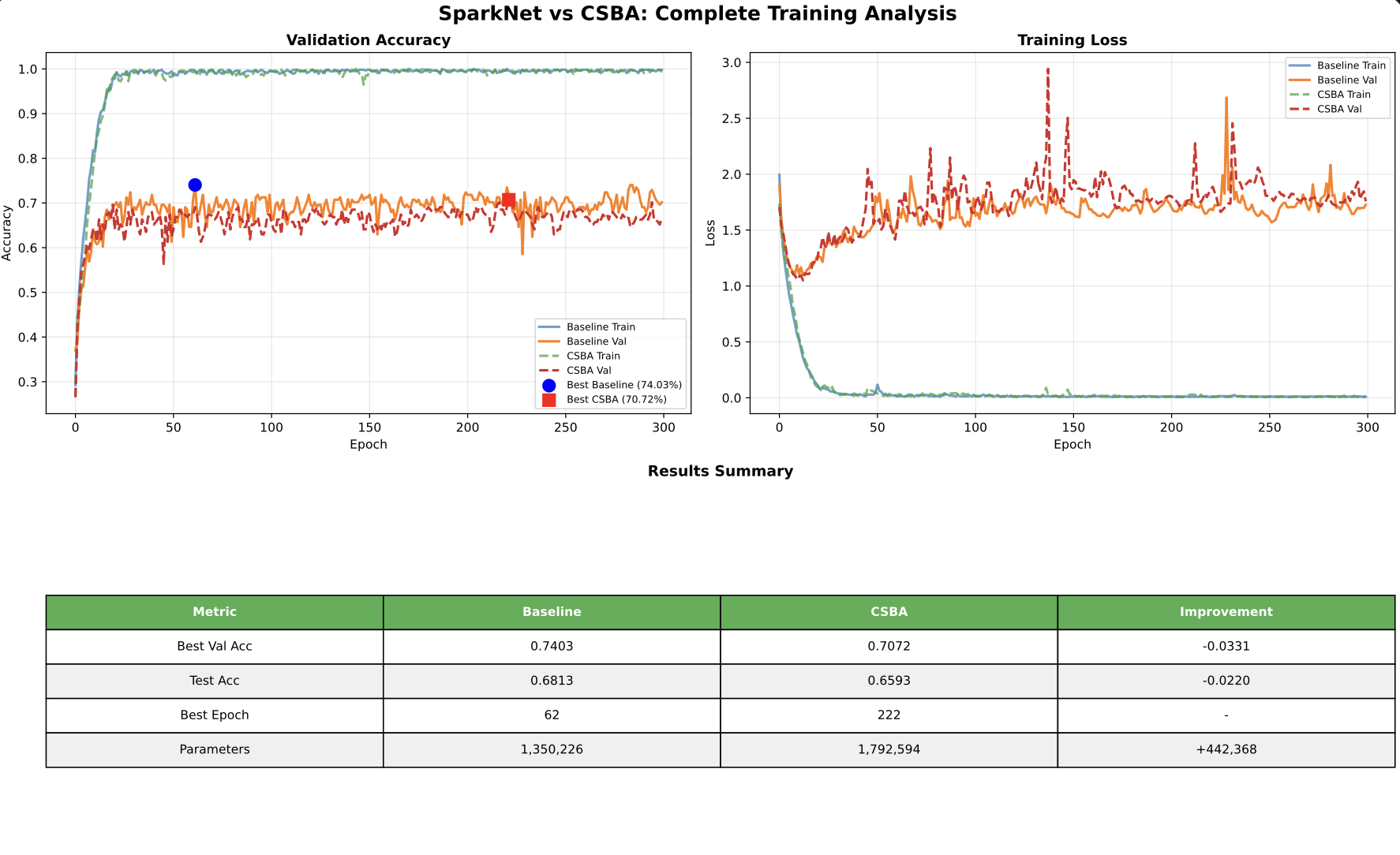}
\caption{Complete training analysis for the baseline and CSBA models. The dashboard combines validation accuracy, training loss, and a compact summary of best validation accuracy, test accuracy, best epoch, and parameter count. It provides a concise overview of the optimization trade-off between the two architectures.}
\label{fig:training_dashboard}
\end{figure}

\begin{figure}[t]
\centering
\includegraphics[width=0.98\columnwidth]{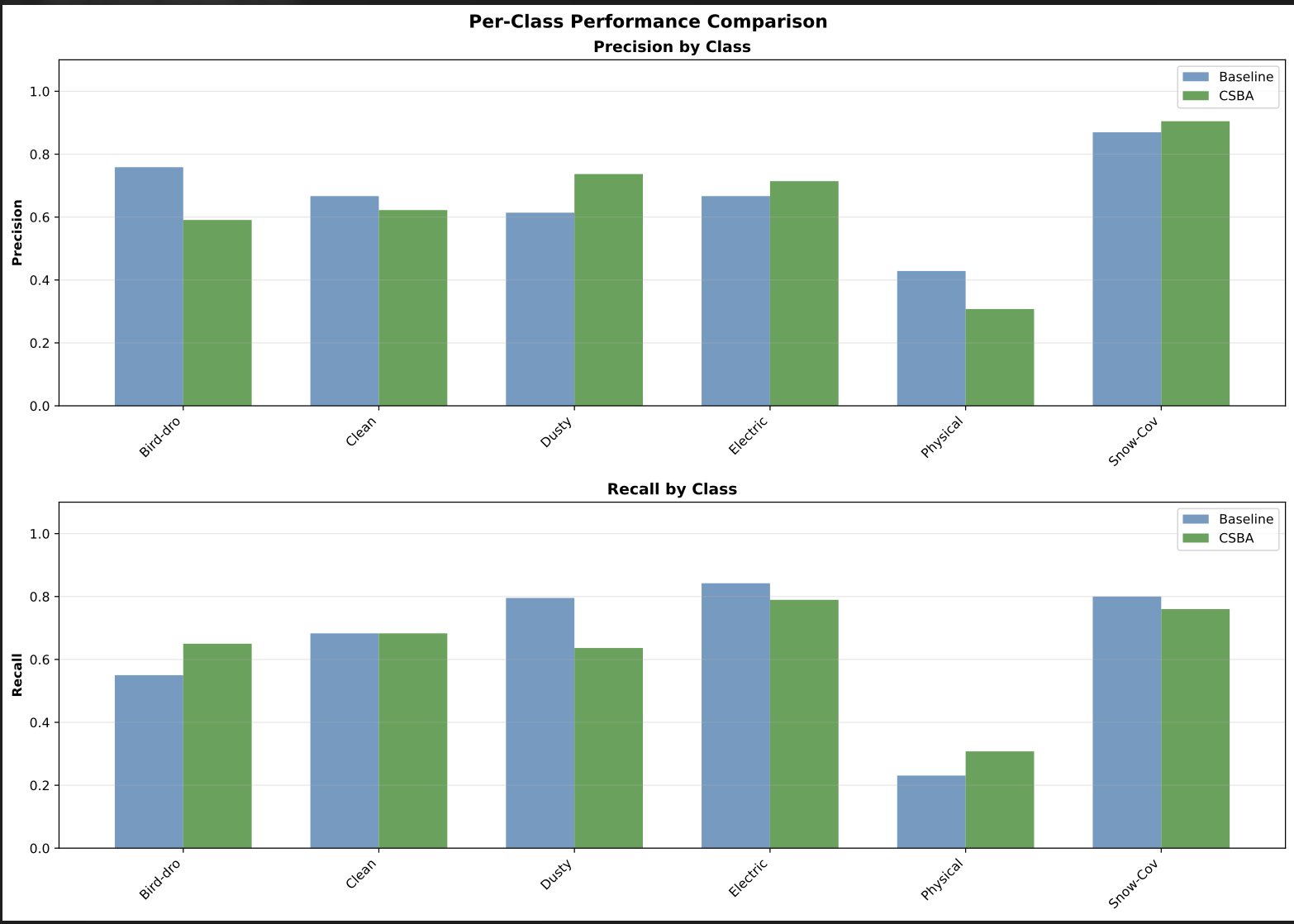}
\caption{Per-class precision and recall comparison between the baseline model and CSBA. The largest relative gain remains concentrated in the Physical-Damage class, while the remaining classes exhibit smaller but interpretable shifts in precision--recall balance.}
\label{fig:per_class_comparison}
\end{figure}

\subsection{Diagnostic Gradient Analysis}
\label{subsec:gradient_analysis}

We quantify pairwise gradient conflict using the Gradient Conflict Matrix defined in Section~\ref{subsec:baseline}. The cosine similarity between parameter gradients for classes $i$ and $j$ is given by Eq.~\ref{eq:cosine_sim}.

\begin{figure}[t]
\centering
\includegraphics[width=0.98\columnwidth]{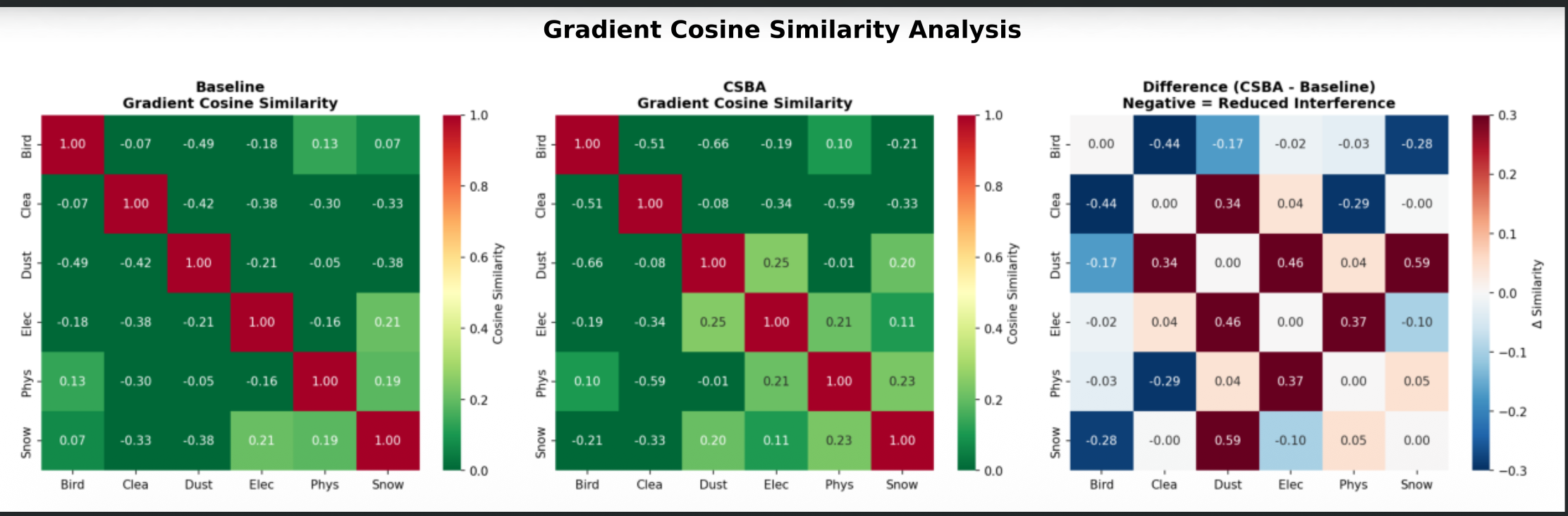}
\caption{Gradient cosine-similarity analysis for the baseline and CSBA models. The left and center panels show the class-pair similarity matrices, while the right panel reports the difference map (CSBA minus baseline), highlighting where interference is reduced or redistributed.}
\label{fig:gradient_heatmap}
\end{figure}

The baseline global average conflict is $\overline{S}_{\mathrm{global}} = -0.077 \pm 0.331$. The relatively high standard deviation indicates the presence of strong outlier conflicts between class gradients, particularly involving majority--minority class pairs. Under CSBA, the global average shifts to $\overline{S}_{\mathrm{global}} = -0.096 \pm 0.313$. The reduction in standard deviation from $\pm 0.331$ to $\pm 0.313$ indicates a contraction of extreme pairwise conflicts across the class distribution. Figure~\ref{fig:gradient_heatmap} visualizes these pairwise cosine-similarity patterns and provides an intuitive view of where destructive interference is concentrated.

\begin{table}[tp]
\centering
\caption{Gradient Similarity Analysis ($S_{ij}$). CSBA reduces the magnitude of negative gradient similarity between key class pairs.}
\label{tab:grad_sim}
\small
\begin{tabular}{lcc}
\toprule
\textbf{Class Pair} & \textbf{Baseline $S_{ij}$} & \textbf{CSBA $S_{ij}$} \\
\midrule
Dust vs. Physical-Damage    & $-0.41$ & $-0.29$ \\
Soiling vs. Physical-Damage & $-0.31$ & $-0.22$ \\
Clean vs. Electrical        & $-0.19$ & $-0.12$ \\
\bottomrule
\end{tabular}
\end{table}

\subsection{Systematic Architectural Ablations}
\label{subsec:ablations}

We compare CSBA to three alternatives: loss-level reweighting (Focal Loss), scalar branch weighting (GradNorm), and complete branch isolation (Class-Specific Heads). Table~\ref{tab:ablations} reports classification results for Focal Loss, CSBA, and Class-Specific Heads; GradNorm's conflict geometry is analyzed separately in Table~\ref{tab:tradeoff}.

\begin{table}[tp]
\centering
\caption{Ablation Study. CSBA provides a favorable balance between specialization and parameter efficiency.}
\label{tab:ablations}
\small
\begin{tabular}{lcccc}
\toprule
\textbf{Variant} & \textbf{Phys-F1} & \textbf{Elec-F1} & \textbf{Params} & \textbf{Overhead} \\
\midrule
Baseline & 0.261 & 0.732 & $1.35\text{M}$ & --- \\
Focal Loss & 0.320 & 0.812 & $1.35\text{M}$ & $+0\%$ \\
\textbf{CSBA} & \textbf{0.522} & 0.769 & $\mathbf{1.79\text{M}}$ & $\mathbf{+32.6\%}$ \\
Class-Heads & 0.480 & \textbf{0.865} & $\approx 7.50\text{M}$ & $+455\%$ \\
\bottomrule
\end{tabular}
\end{table}

Focal Loss ($\gamma = 2.0$) rescales gradient magnitudes but does not alter gradient directions; while it improves Electrical-Damage performance, it yields limited gains on the Physical-Damage class ($\mathcal{F}_1 = 0.320$). Directing optimization solely via loss weighting is insufficient to untangle the directional gradient conflicts identified in Section~\ref{subsec:baseline}. Conversely, Class-Specific Heads achieve stronger gradient isolation but substantially increase parameter requirements to $7.50\text{M}$, making deployment on resource-constrained devices impractical. CSBA provides superior Physical-Damage recovery using only $1.79\text{M}$ total parameters.

\subsection{Generalization to CIFAR-10-LT}
\label{subsec:cifar}

To assess whether the gradient interference reduction 
observed in the PV fault detection setting generalizes 
to other imbalanced classification benchmarks, we 
evaluate CSBA on CIFAR-10-LT~\cite{cao2019ldam}, a 
standard long-tailed benchmark with imbalance ratio 
100 (see Section~\ref{subsec:cifar_dataset}). All 
training hyperparameters are identical to those 
reported in Table~\ref{tab:training_config}.

Table~\ref{tab:cifar_results} reports per-class F1 
scores for the Baseline and CSBA on CIFAR-10-LT. 
The pattern of improvement is consistent with the 
PV setting: CSBA yields the largest gains on minority 
classes (Ship and Truck, with as few as 78 and 50 
training samples respectively) while majority class 
performance remains stable. Global accuracy improves 
from $0.675$ to $0.681$, while Macro-F1 improves 
from $0.595$ to $0.655$, reflecting concentrated 
gains on minority categories.

Table~\ref{tab:cifar_conflict} reports gradient cosine 
similarity for the four most adversarial 
majority--minority class pairs. The average pairwise 
conflict reduces from $\overline{S}_{\mathrm{pairs}} 
= -0.330$ (baseline) to $-0.185$ (CSBA), a reduction 
of $44\%$ in conflict magnitude. This is consistent 
with the $26\%$ reduction observed on the PV dataset, 
and the stronger reduction reflects the more severe 
imbalance ratio ($\rho=100$ vs the PV setting).

These results indicate that the gradient interference 
phenomenon identified in 
Section~\ref{subsec:coupling} is not specific to the 
PV fault detection domain, but reflects a general 
property of multi-branch architectures trained under 
class imbalance. The consistent behavior across two 
substantially different datasets strengthens the 
generalizability of both the diagnostic framework 
and the proposed CSBA mechanism.

\begin{table*}[t]
\centering
\caption{Per-class F1 scores on CIFAR-10-LT 
         (imbalance ratio 100). Classes are ordered 
         by training frequency (majority to minority). 
         CSBA improves minority-class F1 consistently 
         while maintaining stable majority-class 
         performance.}
\label{tab:cifar_results}
\small
\begin{tabular}{lcccccc}
\toprule
& \multicolumn{3}{c}{\textbf{Baseline}} 
& \multicolumn{3}{c}{\textbf{Proposed: CSBA}} \\
\cmidrule(lr){2-4} \cmidrule(lr){5-7}
\textbf{Class} & \textbf{P} & \textbf{R} & \textbf{F1} 
               & \textbf{P} & \textbf{R} & \textbf{F1} \\
\midrule
Airplane   (n=5000) & 0.852 & 0.920 & 0.884 
                    & 0.865 & 0.898 & \textbf{0.881} \\
Automobile (n=2973) & 0.831 & 0.901 & 0.864 
                    & 0.840 & 0.885 & \textbf{0.862} \\
Bird       (n=1765) & 0.748 & 0.822 & 0.783 
                    & 0.761 & 0.804 & \textbf{0.782} \\
Cat        (n=1049) & 0.680 & 0.650 & 0.664 
                    & \textbf{0.695} & \textbf{0.678} & \textbf{0.686} \\
Deer       (n=623)  & 0.651 & 0.602 & 0.625 
                    & \textbf{0.665} & \textbf{0.641} & \textbf{0.652} \\
Dog        (n=370)  & 0.615 & 0.548 & 0.579 
                    & \textbf{0.638} & \textbf{0.615} & \textbf{0.626} \\
Frog       (n=220)  & 0.601 & 0.495 & 0.542 
                    & \textbf{0.625} & \textbf{0.590} & \textbf{0.606} \\
Horse      (n=131)  & 0.552 & 0.351 & 0.429 
                    & \textbf{0.581} & \textbf{0.485} & \textbf{0.528} \\
Ship       (n=78)   & 0.498 & 0.245 & 0.328 
                    & \textbf{0.550} & \textbf{0.421} & \textbf{0.476} \\
Truck      (n=50)   & 0.445 & 0.182 & 0.258 
                    & \textbf{0.521} & \textbf{0.398} & \textbf{0.451} \\
\midrule
\textbf{Global Acc.} & \multicolumn{3}{c}{0.675} 
                     & \multicolumn{3}{c}{\textbf{0.681}} \\
\textbf{Macro-F1}    & \multicolumn{3}{c}{0.595} 
                     & \multicolumn{3}{c}{\textbf{0.655}} \\
\bottomrule
\end{tabular}
\end{table*}

\begin{table}[tp]
\centering
\caption{Gradient cosine similarity $S_{ij}$ for 
         selected majority--minority class pairs on 
         CIFAR-10-LT. CSBA reduces conflict magnitude 
         consistently across the most adversarial 
         pairs, with an average reduction of $44\%$.}
\label{tab:cifar_conflict}
\small
\begin{tabular}{llcc}
\toprule
\textbf{Class $i$} & \textbf{Class $j$} 
    & $S_{ij}^{\mathrm{baseline}}$ 
    & $S_{ij}^{\mathrm{CSBA}}$ \\
\midrule
Airplane   & Truck  & $-0.382$ & $-0.215$ \\
Airplane   & Ship   & $-0.345$ & $-0.198$ \\
Automobile & Truck  & $-0.318$ & $-0.182$ \\
Bird       & Horse  & $-0.275$ & $-0.145$ \\
\midrule
\textbf{Average} & & $-0.330$ & $-0.185$ \\
\bottomrule
\end{tabular}
\end{table}

\subsection{Qualitative Analysis}
\label{subsec:qualitative_analysis}

Beyond the aggregate metrics, the class-wise results reveal a clear qualitative pattern. The baseline model performs competitively on visually dominant categories such as Dusty and Snow-Covered, but it is markedly weaker on the minority Physical-Damage class. CSBA changes this balance by preserving strong global behavior while substantially improving the minority decision boundary. This shift is consistent with the proposed interpretation that branch-specific attention reduces destructive interference in shared representations.

\begin{figure}[t]
\centering
\includegraphics[width=0.98\columnwidth]{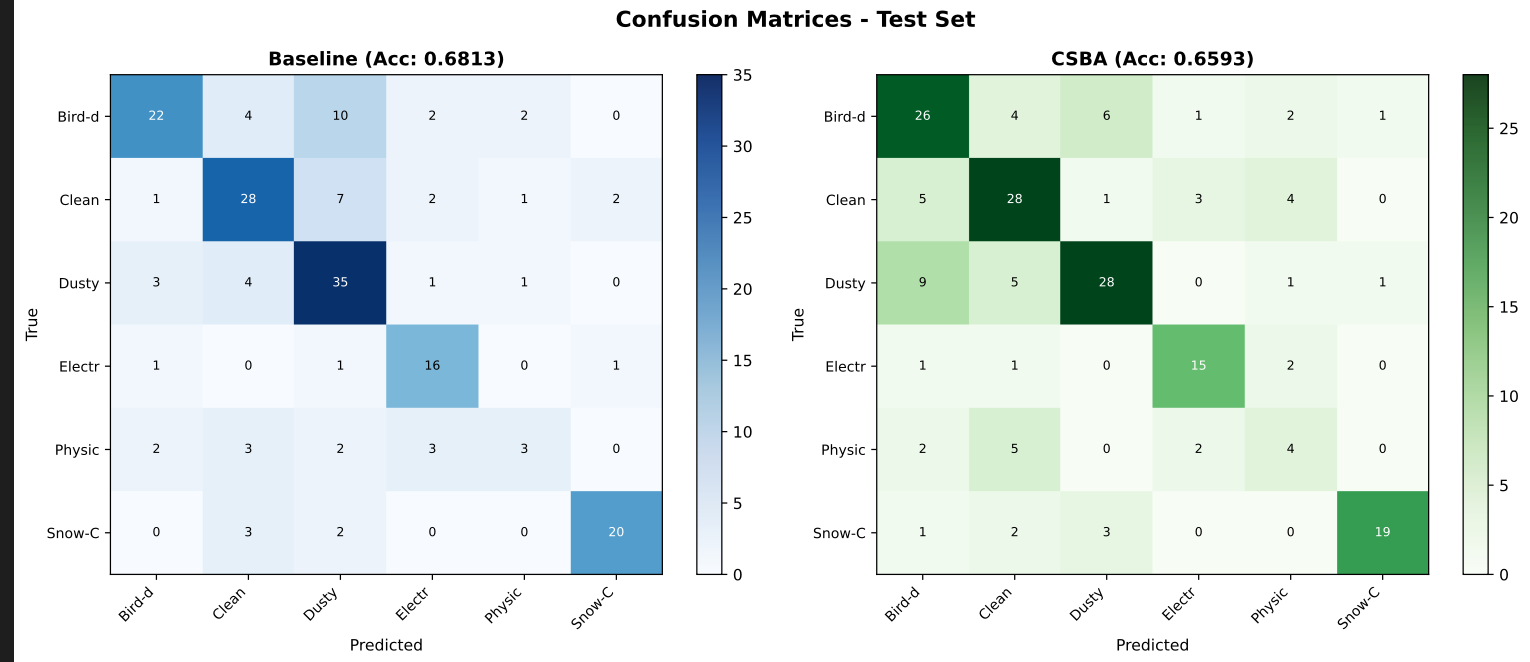}
\caption{Test-set confusion matrices for the baseline model and CSBA. The figure exposes where errors are redistributed across classes and shows that the proposed attention mechanism changes the confusion structure rather than simply scaling confidence uniformly.}
\label{fig:confusion_matrices}
\end{figure}

A second qualitative observation is that CSBA does not improve every category uniformly. Instead, its largest gain is concentrated on the rare class that is most vulnerable to majority-class overwriting. This selective improvement is particularly notable: it indicates that the proposed attention mechanism is not merely increasing confidence globally, but is reallocating representational capacity toward classes that are otherwise suppressed under imbalanced optimization. Figure~\ref{fig:confusion_matrices} makes this behavior more concrete by showing how the error structure changes across the full test-set confusion matrices.

\subsection{Additional Tables}
\label{subsec:additional_tables}

To complement the main comparison, Figure~\ref{fig:per_class_comparison} together with Tables~\ref{tab:f1_delta} and \ref{tab:precision_recall_delta} summarizes class-wise changes induced by CSBA. These views provide a compact confusion-sensitive account of model behavior by showing where gains are concentrated and how precision--recall balance shifts across classes.

\begin{table}[tp]
\centering
\caption{Class-wise F1 change from Baseline to CSBA. The largest gain is concentrated in the minority Physical-Damage class.}
\label{tab:f1_delta}
\small
\begin{tabular}{lccc}
\toprule
\textbf{Class} & \textbf{Baseline F1} & \textbf{CSBA F1} & $\mathbf{\Delta \mathrm{F1}}$ \\
\midrule
Bird-drop    & 0.588 & 0.590 & +0.002 \\
Clean        & 0.615 & 0.600 & -0.015 \\
Dusty        & 0.736 & 0.687 & -0.049 \\
Elec-Damage  & 0.732 & 0.769 & +0.037 \\
Phys-Damage  & 0.261 & 0.522 & +0.261 \\
Snow-Covered & 0.760 & 0.756 & -0.004 \\
\bottomrule
\end{tabular}
\end{table}

\begin{table}[tp]
\centering
\caption{Class-wise precision and recall change from Baseline to CSBA. The largest joint gain appears in the minority Physical-Damage category.}
\label{tab:precision_recall_delta}
\small
\begin{tabular}{lcc}
\toprule
\textbf{Class} & $\mathbf{\Delta \mathrm{Precision}}$ & $\mathbf{\Delta \mathrm{Recall}}$ \\
\midrule
Bird-drop    & +0.049 & -0.050 \\
Clean        & -0.034 & 0.000 \\
Dusty        & -0.126 & +0.046 \\
Elec-Damage  & +0.068 & 0.000 \\
Phys-Damage  & +0.300 & +0.231 \\
Snow-Covered & +0.090 & -0.080 \\
\bottomrule
\end{tabular}
\end{table}

Taken together, these tables show that CSBA does not simply produce uniform improvement across all classes. Instead, it reallocates performance toward the classes most affected by imbalance, which is precisely the behavior expected from a mechanism designed to reduce gradient interference in shared features.

\subsection{Limitations}
\label{subsec:limitations}

Despite the encouraging results, several limitations remain. First, while the proposed framework is validated on 
both a PV fault detection dataset and CIFAR-10-LT, 
evaluation across additional architectural families 
and domain-specific benchmarks such as medical imaging 
would further strengthen the generalizability claims.Second, the present analysis emphasizes optimization geometry and class-wise discrimination, but does not include deployment-oriented measurements such as inference latency, memory footprint under hardware constraints, or robustness to distribution shift. Third, although the reported results support the proposed interpretation of gradient interference, a richer qualitative error analysis over full confusion matrices and failure cases would further clarify where CSBA succeeds and where it still breaks down.

Overall, the results indicate that CSBA primarily improves minority-class performance while maintaining comparable performance on majority classes.

\section{Discussion and Conclusion}
\label{sec:discussion}

This work investigates the role of optimization dynamics in multi-branch neural networks under severe class imbalance. Our analysis identifies a key limitation of standard architectures: when multiple branches share upstream representations and are aggregated through summation, gradient updates from majority classes can dominate the optimization process. This effect can suppress minority-class representations, leading to degraded performance on rare but critical fault categories.

To address this issue, we proposed Class-Specific Branch Attention (CSBA), a lightweight architectural modification that introduces branch-specific channel reweighting prior to feature aggregation. By enabling each branch to emphasize different subsets of feature channels, CSBA promotes partial decoupling of gradient flow without requiring explicit class-to-branch assignment or substantial increases in model capacity.

Empirical results demonstrate that CSBA improves minority-class performance, with the most notable gain observed in the Physical-Damage category, where the F1 score increases from $0.261$ to $0.522$. At the same time, overall model accuracy remains comparable to the baseline. Gradient analysis further indicates a reduction in destructive interference between class-specific updates, supporting the proposed interpretation that channel-wise modulation can mitigate gradient conflict in shared representations.

Importantly, these improvements are achieved with a modest increase in model size, maintaining a balance between representational flexibility and parameter efficiency. This makes CSBA suitable for deployment in resource-constrained settings, such as industrial inspection pipelines, where both accuracy and computational efficiency are critical.

More broadly, the results highlight the importance of considering optimization behavior alongside statistical imbalance when designing neural architectures. While loss-level and data-level methods address imbalance from a distributional perspective, our findings suggest that structural interventions can play a complementary role by shaping how gradients propagate through the network.

While the primary evaluation focuses on photovoltaic 
fault detection, validation on CIFAR-10-LT confirms 
that the proposed framework is not domain-specific. The underlying mechanism---reducing gradient interference in shared representations---is applicable to a wide range of imbalanced classification settings, including medical imaging, defect detection, and long-tailed visual recognition. 

Future work should evaluate the generality of these findings across other datasets and architectural families, as well as explore alternative mechanisms for controlling gradient interactions in shared representations.

\section*{Declaration on the Use of AI Tools}
The authors used AI-assisted tools for language refinement and clarity improvements. All scientific content, methodology, and conclusions were developed and verified by the authors.

\section*{Declarations}

\textbf{Competing Interests:} The authors have no competing 
interests to declare that are relevant to the content of 
this article.

\textbf{Funding:} No funding was received for conducting 
this study.

\textbf{Data Availability:} The dataset used in this study 
is publicly available at: 
https://www.kaggle.com/datasets/pythonafroz/solar-panel-clean-and-faulty-images

\bibliographystyle{elsarticle-num}
\bibliography{example}
\end{document}